\documentclass[]{fairmeta}

\usepackage[utf8]{inputenc}
\usepackage{tikz}
\usetikzlibrary{arrows.meta,positioning,calc,fit,backgrounds}
\usepackage{pgfplots}
\pgfplotsset{compat=1.18}
\usepackage{fontawesome5}
\usepackage{amsmath}
\usepackage{amssymb}
\usepackage[ruled,linesnumbered]{algorithm2e}
\usepackage{fvextra}
\usepackage{wrapfig}

\newtcolorbox{promptbox}[1]{%
  breakable, enhanced, sharp corners, boxrule=0.5pt,
  colframe=black!55, colback=black!2,
  colbacktitle=black!8, coltitle=black,
  fonttitle=\bfseries\small, title={#1},
  left=4pt, right=4pt, top=3pt, bottom=3pt,
  attach boxed title to top left={yshift=-0.1mm, xshift=4pt},
  boxed title style={sharp corners, boxrule=0.5pt, colframe=black!55},
}

\title{Training Small Language Models for Transferable Tool Selection}

\author[1, 2,*]{Zhaoyi Zhou}
\author[1]{Akshayaa Magesh}
\author[1]{Jeongyeol Kwon}
\author[1]{Artem Kuriksha}
\author[1]{Daniel Jiang}
\author[2]{Andrea Zanette}
\author[1]{Ayush Jain}

\affiliation[1]{Meta}
\affiliation[2]{Carnegie Mellon University}

\contribution[*]{Work done at Meta}

\abstract{
As LLM agents operate over corpora of more than ten thousand tools, selecting
the right tool for a complex multi-step user instruction becomes a central bottleneck. Tool descriptions cannot all fit in context, and relevance-based retrievers only rank tools
without knowing which are actually needed. Recent work instead leverages proprietary commercial LLMs like Claude or ChatGPT to
drive a retriever, decomposing the instruction and selecting over
retrieved candidates. We show that a small, open model can be \emph{trained} via reinforcement learning to
perform the same functionality. On ToolBench, a Qwen3-8B model post-trained with our recipe outperforms Llama-4-Maverick (402B). In addition, the trained model transfers directly to different retrievers and generalizes to new tool corpora in different domains.
}

\date{October 22, 2026}
\correspondence{Zhaoyi Zhou at \email{zhaoyiz@andrew.cmu.edu}}

\begin{document}

\maketitle

\section{Introduction}
\label{sec:intro}

\begin{figure*}[t]
\centering
\includegraphics[width=\textwidth]{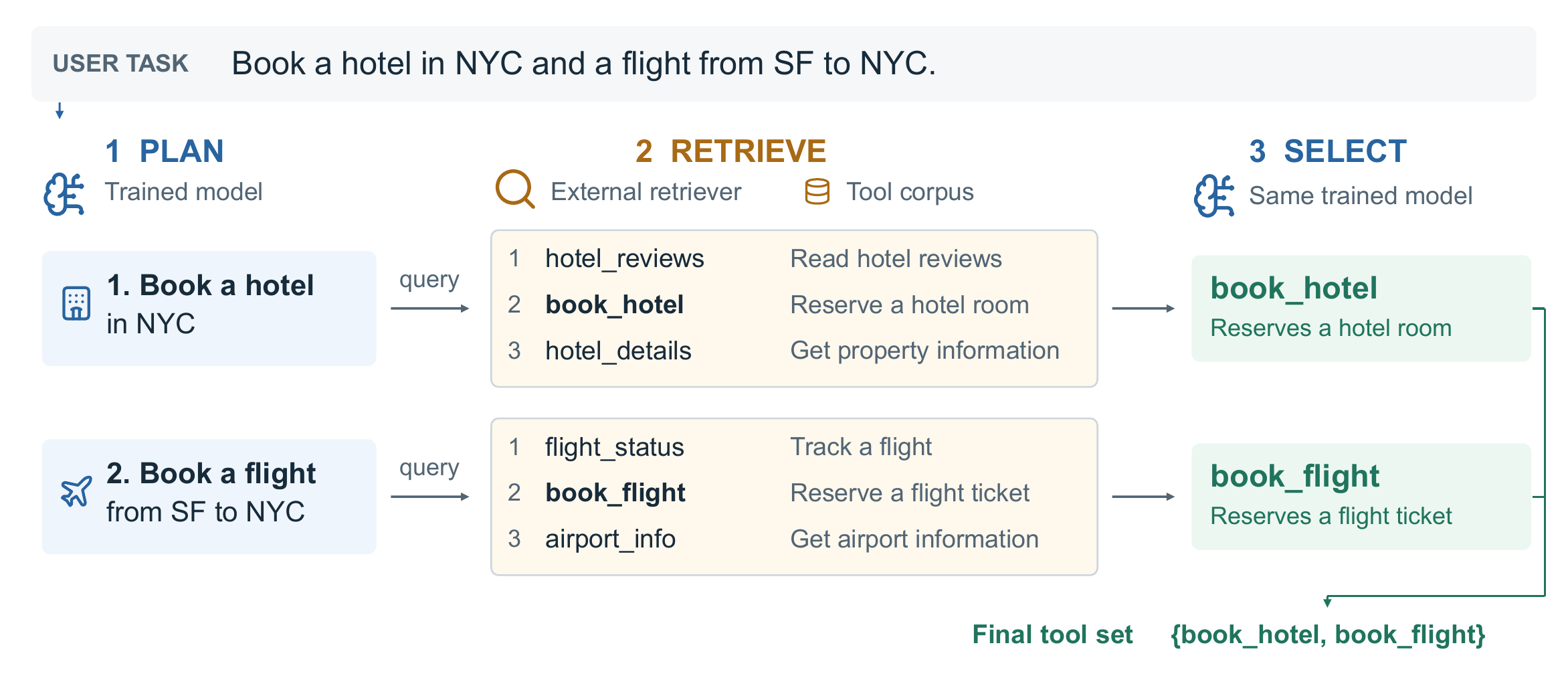}
\caption{Illustration of the \textsc{Plan}--\textsc{Retrieve}--\textsc{Select} tool selection workflow. The same trained model decomposes the request and selects a tool for each substep. Each substep issues one query to an external retriever, which filters candidates by score and retains at most $K$ for the \textsc{Select} stage.}
\label{fig:teaser}
\end{figure*}

Large language model (LLM) agents increasingly solve user tasks by
invoking external tools
(or APIs)\footnote{ToolBench~\citep{qin_toolllm_2024} uses ``tool'' for a RapidAPI service that groups several APIs. In this paper the unit being selected is always an individual API, which we call a tool.}
e.g., sending an email, booking a flight, or fetching data
from the web, rather than producing an answer from the model's learned
knowledge alone. As agents are applied to real software ecosystems, the number
of callable tools has grown to well over ten thousand in some public tool
corpora~\citep{qin_toolllm_2024,shiRetrievalModelsArent2025}, and enterprises
add proprietary tools for internal use. At this scale, the model's context
overflows if every tool description is placed in it, so selecting the right tool
becomes a bottleneck that precedes the actual execution.

The classical solution is a \emph{retriever}, which scores each tool by its
similarity to the user instruction and returns a small set of high-scoring
candidates for the agent to choose from.
Retrievers come in two broad families. Lexical retrievers such as BM25
(Best Matching 25)~\citep{robertson_bm25_2009} score a tool by how many
query terms appear in its documentation, weighting rarer terms more heavily,
and apply to any corpus without training. Dense retrievers such as
COLT~\citep{quCompletenessOrientedToolRetrieval2024} instead embed the
instruction and each tool description into a shared vector space with an
encoder trained on the target corpus, and rank tools by embedding similarity.
LLMs have also been introduced to assist a fixed retriever by rewriting
queries~\citep{chen_re-invoke_2024,kachuee_improving_2025}, expanding
documents~\citep{lu_tools_2026,senthil_cohyde_2026,wu_gretel_2025}, or
re-ranking candidates~\citep{zheng_toolrerank_2024}. However, a retriever can
only assign a relevance-based ranking over tools; it does not know which of
them the task actually requires. This is especially problematic when the user
instruction contains multiple substeps and requires multiple tools.

To address this, recent work places the LLM at the center of the selection
workflow and uses the retriever as a \emph{meta-tool} invoked
on demand. Instead of consuming a single ranked list produced in one pass,
the LLM drives retrieval across turns in a ReAct-style
loop~\citep{braunschweilerToolReAGtToolRetrieval2025,lumer_scalemcp_2025,lumer_memtool_2025}:
a \textsc{Plan} stage decomposes the user instruction into substeps, a
\textsc{Retrieve} stage fetches a set of relevant candidates, and a \textsc{Select}
stage chooses the tools it judges necessary, repeating until the instruction
is covered. Because the model, not the ranking function, decides which retrieved tools are required, this design directly handles the multi-substep, multi-tool instructions that a relevance ranking alone cannot address.
Existing work performs tool selection with proprietary commercial models such
as Claude or ChatGPT under careful prompt tuning. Since tool selection is a
recurring \emph{sub-routine} of the agentic tool-calling pipeline, we ask
whether a \textbf{small} model can perform the task.

While an off-the-shelf small model performs poorly on its own,
fine-tuning specifically for tool selection is a viable solution. However, this raises the key technical challenge of
\textbf{transferability}:
\begin{itemize}
  \item \textbf{Transferability across tool domains.} A tool-selection model
  trained on one domain of tools should be usable directly on another, so that
  a separate model need not be trained for every tool corpus.
  \item \textbf{Transferability across retrievers.} Some tools carry specific
  tool names and thus favor exact string matching, whereas other tools are better found by semantic matching, so the preferred retriever
  depends on the underlying tool corpus. We want the model to work under a new
  retriever without retraining.
\end{itemize}

\begin{figure*}[t]
\centering
\includegraphics[width=\textwidth]{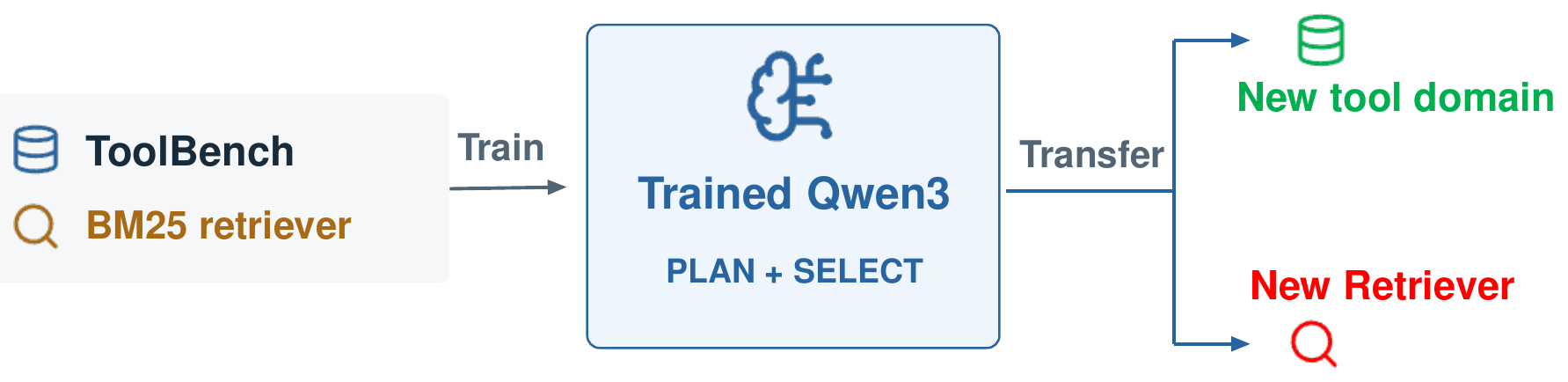}
\caption{Once trained, our model directly transfers to other tool corpora in new domains and is compatible with new retrievers, without additional model training.}
\label{fig:transfer-teaser}
\end{figure*}

These challenges are tractable under the studied framework because tool selection is built on the generalizable skills
of the underlying LLM: decomposing the user instruction and matching
a tool description to it. The LLM needs no domain-specific knowledge, and the
workflow stays invariant to different retrievers. We therefore fine-tune Qwen3 models~\citep{yang2025qwen3} with reinforcement learning to acquire these skills, and transfer them
to new domains and retrievers (Figure~\ref{fig:transfer-teaser}).

Our main contributions are as follows:
\begin{itemize}
\item \textbf{Training a small LLM for tool selection.} We fine-tune
  off-the-shelf Qwen3 models (1.7B, 4B, and 8B) with RL. On ToolBench, their
  performance exceeds that of Llama-4-Maverick (402B)~\citep{meta2025llama4} with prompt tuning.
\item \textbf{Domain transferability.} The trained LLM transfers zero-shot
  across tool corpora: trained on ToolBench G3, it selects tools on G2,
  Gorilla, CRAFT, and TaskBench with no additional training.
\item \textbf{Retriever transferability.} A model trained over BM25 drives a
  domain-specialized dense retriever (COLT) at inference without retraining,
  retaining its learned selection behavior.
\end{itemize}

\section{Related Work}
\label{sec:related}

\paragraph{Tool retrievers.} Conventional approaches treat tool selection as a
single-shot retrieval over tool descriptions. ToolBench trains a Sentence-BERT
retriever~\citep{qin_toolllm_2024}; ToolRet benchmarks tool retrieval and shows
that general-purpose IR encoders transfer poorly to
tools~\citep{shiRetrievalModelsArent2025}; COLT adds a query--tool--scene graph
to optimize for completeness~\citep{quCompletenessOrientedToolRetrieval2024};
ToolGen represents each tool as a vocabulary token~\citep{wang_toolgen_2025};
ToolRerank adds hierarchy-aware reranking~\citep{zheng_toolrerank_2024}; and
\citet{gao_task-aligned_2025} size the returned set per task instead of a fixed
top-$k$. Beyond retrieval, MetaTool~\citep{huang_metatool_2024} benchmarks
whether a model knows when and which tool to use.

\paragraph{LLM-assisted retrieval.} An LLM can also improve a fixed retriever.
It can expand tool documents offline~\citep{lu_tools_2026,senthil_cohyde_2026,wu_gretel_2025,lumer_toolshed_2024},
rewrite the query for single-pass retrieval~\citep{chen_re-invoke_2024,kachuee_improving_2025}, or operate the
retriever across turns: iterative LLM
feedback~\citep{xu_enhancing_2024}, mid-conversation
re-querying~\citep{lumer_scalemcp_2025}, short-term tool
memory~\citep{lumer_memtool_2025}, and ReAct-style retrieval with one query per
decomposed substep~\citep{braunschweilerToolReAGtToolRetrieval2025}. Our
workflow follows the multi-turn design, but trains a small model instead of prompting a large one.

\paragraph{Action generalization in RL.} Standard RL assumes a fixed action set,
whereas our actions are tools from a corpus that could change between training and
deployment. Prior work handles large or changing action sets by learning action
representations~\citep{dulac-arnold_deep_2015,chandak_learning_2019,chandak_lifelong_2020},
and generalizes zero-shot to unseen actions by inferring their representations
from action information acquired separately from the
task~\citep{jain_generalization_2020,jain_know_2022}. Our \textsc{Select} stage is such a
policy over a varying action set of the retrieved tools, with each
represented by its documentation. The LLM supplies these representations implicitly without a separately trained action encoder.

\paragraph{RL-trained LLM agents.} DeepAgent~\citep{li_deepagent_2026} trains an end-to-end deep reasoning agent covering tool discovery and execution. \citet{zhang_data-efficient_2024} derive DPO preferences for single-shot query rewriting from a frozen retriever's rankings, and ToolRL~\citep{qian_toolrl_2025} studies reward design for tool learning. None of them study transferability in tool selection. Beyond tool selection, RL has also been applied to other search problems: open-domain QA~\citep{jin_search-r1_2025,song_r1-searcher_2025,chen_research_2025,sun_zerosearch_2026}, query generation~\citep{jiang_deepretrieval_2025}, web agents~\citep{zheng_deepresearcher_2025,li_webthinker_2025,qiWebRLTrainingLLM2025}, and retriever training from agent trajectories~\citep{zhou_learning_2026}.

\section{Method}
\label{sec:method}

Given a user instruction $x$ and a tool corpus $\mathcal{T}$ too large to fit
in the model's context, tool selection asks for the subset
$T \subseteq \mathcal{T}$ of tools that $x$ requires. Rather than training a
corpus-specific dense retriever, we let a single small model $\pi$
\emph{operate} a lightweight, corpus-agnostic retriever $\mathcal{R}$ to find
these tools. We first define the workflow through which $\pi$ drives
$\mathcal{R}$ (\S\ref{sec:workflow}), and then describe how $\pi$ is trained:
supervised fine-tuning (\S\ref{sec:sft}) followed by reinforcement learning
(\S\ref{sec:rl}). Unless otherwise stated, $\mathcal{R}$ is
BM25~\citep{robertson_bm25_2009}, which requires no training and applies to
any corpus. The transfer to new tool domains in \S\ref{sec:experiments}
therefore comes from the trained model rather than from a retriever fitted to
the target corpus.

\subsection{The tool-selection workflow}
\label{sec:workflow}

The LLM drives the retriever through a three-stage workflow
(Figure~\ref{fig:teaser} and Algorithm~\ref{alg:workflow}): a \textsc{Plan}
stage decomposes the user instruction into substeps, a \textsc{Retrieve} stage
fetches relevant candidates, and a \textsc{Select} stage chooses the tools it
judges necessary, repeating until the instruction is covered.

\paragraph{\textsc{Plan}.} A user instruction often contains several requests. For instance, ``book a flight from San Francisco to New York City and find a hotel in New York City'' naturally splits into two. Because the retriever scores tools against the whole query, the terms of one request interfere with retrieval for the other. The model therefore first decomposes the instruction into an ordered list of substeps, each a short natural-language description of a single required capability.

\paragraph{\textsc{Retrieve}.} For each substep, the retriever scores all tool descriptions against the substep text in a single query. We keep tools scoring above a threshold $\tau$ and truncate them to the top $K$, which filters out irrelevant tools and bounds the context loaded at the \textsc{Select} stage. Unless otherwise stated, $\tau=3$ and $K=15$. We train with BM25, but any retriever that scores tools against a substep, such as COLT~\citep{quCompletenessOrientedToolRetrieval2024}, a dense retriever trained on the ToolBench corpus, can be substituted at test time without retraining.

\paragraph{\textsc{Select}.} The descriptions of the candidate tools are loaded
into the LLM's context, and the LLM selects the tool that best fulfills the
substep. This assumes that each substep can be solved by exactly one tool and
that substeps are independent, i.e., no substep needs the output of another. The
selected tools are accumulated across substeps, and their union is returned as
the final tool set. The same model $\pi$ serves both LLM stages under two
different prompts (Appendix~\ref{app:prompts}). We write $\pi_{\textsc{Plan}}$
and $\pi_{\textsc{Select}}$ only to indicate which prompt is used. Each LLM call
is stateless and sees only the user instruction (\textsc{Plan}) or the current
substep with its $\leq K$ candidates (\textsc{Select}), so the per-call context
is bounded regardless of corpus size or plan length.

\begin{figure}[t]
\centering
\begin{minipage}{0.75\textwidth}
\begin{algorithm}[H]
\DontPrintSemicolon
\small
\SetKwInput{Input}{Input}\SetKwInput{Output}{Output}
\SetKwFunction{Retr}{$\mathcal{R}$}
\SetKwFunction{PiPlan}{$\pi_{\textsc{Plan}}$}
\SetKwFunction{PiSelect}{$\pi_{\textsc{Select}}$}
\SetKw{Continue}{continue}
\Input{User instruction $x$; model $\pi$; retriever $\mathcal{R}$; score threshold $\tau$; candidate cap $K$}
\Output{Predicted tool set $T$}
\tcp{\textsc{Plan}}
Model $\pi_{\textsc{Plan}}$ decomposes $x$ into substep texts $s_1,s_2,\cdots, s_n$; $T \gets \emptyset$\;
\ForEach{substep text $s_i$}{
  \tcp{\textsc{Retrieve}}
  Retriever $\mathcal{R}$ scores every tool by $s_i$.\;
  $C_i \gets $ Set of tools with score above $\tau$. \;
  Truncate $C_i$ to top-$K$ tools according to score.\;
  \lIf{$C_i = \emptyset$}{\Continue}
  \tcp{\textsc{Select}}
  Model $\pi_{\textsc{Select}}$ selects one tool \textit{id} from $C_i$ given $s_i$. \;
  $T \gets T \cup (\{\textit{id}\} \cap C_i)$.\;
}
\Return $T$\;
\caption{Tool-selection workflow.}
\label{alg:workflow}
\end{algorithm}
\end{minipage}
\end{figure}

\subsection{Data Curation for Supervised Fine-Tuning}
\label{sec:sft}

Reinforcement learning from an off-the-shelf base model is slow to start as the
base model rarely completes the workflow successfully, so we warm-start the policy with
supervised fine-tuning (SFT) on synthetic trajectories before the RL stage
(\S\ref{sec:rl}). We use Llama-4-Maverick as the teacher model to create the training examples.

\paragraph{Gold tool filtering.} We assume access to a reference gold tool set for each query in the training data, but this set can be noisy. For instance, ToolBench~\citep{qin_toolllm_2024} collects gold tools from a reference solution trajectory, so some gold tools are not determined by anything the user wrote, and others are redundant near-duplicates. Training on such tools would teach the
agent to select tools that the instruction does not call for. Therefore, our filter keeps a \emph{(query, gold tool)} pair only if the tool can be identified from the query alone. A deterministic first pass keeps any pair whose query and tool documentation share a content word. The remaining pairs go to the teacher model, which keeps only those whose query contains a paraphrase or synonym of the tool's function. The resulting curated gold set is used for both training
and evaluation.

\paragraph{Plan decomposition.} Given the surviving gold tools, the teacher model decomposes each user instruction into substeps, matching each substep to one of them. It may emit fewer substeps than gold tools, because some gold tools, though relevant, are unnecessary for the instruction. These unused gold tools are removed from the gold set as an additional round of filtering.

\paragraph{Reasoning generation.} The generated substeps and the gold tools
determine the tool-selection trajectory, but we still generate reasoning text so
that the model can more easily learn how to act in the \textsc{Plan} and
\textsc{Select} stages. For \textsc{Plan}, given the user instruction and the generated substeps, the teacher model writes the reasoning that justifies the decomposition. We do not reuse the teacher's reasoning from the decomposition step itself, because the teacher saw the gold tools there and its reasoning would leak them. For \textsc{Select}, the teacher model writes the reasoning that justifies picking the gold tool among the candidates retrieved for the generated substep. A validation step flags any reasoning that names a tool outside the candidate set or leaks candidate indices, and triggers regeneration when a violation occurs.

\paragraph{Fine-tuning.} We collect the reasoning and actions generated during
data curation in JSON format and form \emph{(prompt, reasoning + action)} pairs
for all rounds of LLM generation (prompt templates in
Appendix~\ref{app:prompts}). We then apply standard SFT so that the model learns
the basic thinking pattern and output format for both the \textsc{Plan} and
\textsc{Select} stages, preparing it for RL (\S\ref{sec:rl}).

\subsection{Reinforcement Learning}
\label{sec:rl}

Starting from the SFT checkpoint, we further train the model with RL for better performance (\S\ref{sec:experiments}). For training stability,
we decouple \textsc{Plan} and \textsc{Select} into separate stages: when training
\textsc{Select}, we use the high-quality substeps generated by the teacher model
instead of the model's own plans. Each stage is then a single stateless LLM call
with its own reward, so we can train it with standard
GRPO~\citep{shao_deepseekmath_2024}. The remaining question is how to design the
\textsc{Plan} and \textsc{Select} rewards without additional human annotation.

\paragraph{\textsc{Plan} reward.}
\label{sec:rl-rewards}
The \textsc{Plan} stage is rewarded based on substep text quality, which is reflected by the candidate set retrieved with this text. A plan is good when every gold tool is retrieved by one substep at a high rank
and there are no redundant substeps. Therefore, we treat the substeps as a set and measure its precision and recall against the gold tools.

Suppose the user instruction has $N$ gold tools, and the planning stage emits $n$ substeps. Running
substep $i$ through the retriever yields a
ranked candidate list $A_i$. We try to match every list $A_i$ with a gold tool. We walk the substeps $i$ in
order and scan each $A_i$ in descending order of score. The first gold tool that has not been matched previously is matched to $A_i$, and we move on to the next substep. If no such gold tool is found, $A_i$ is unmatched.

Let $n_{\text{cov}}$ be the number of matched
substeps. Then \emph{precision} is the fraction of substeps that are matched:
\begin{equation*}
\mathrm{precision} = \frac{n_{\text{cov}}}{n}.
\end{equation*}
Precision penalizes unmatched substeps, encouraging every substep to match a unique tool.

Let $i_1,i_2, \cdots, i_{n_{\text{cov}}}$ be the substeps that are matched with a gold tool, and the matched gold tool appears in rank $\rho_j$ for substep $i_j$. \emph{Recall} is the fraction of gold tools covered, graded by rank:
\begin{equation*}
\mathrm{recall} = \frac{1}{N}\sum_{j=1}^{n_{\text{cov}}}\frac{1}{\rho_{j}}.
\end{equation*}
Recall encourages every gold tool to be matched at a higher rank, i.e., the retriever is more confident with the gold tool.

The planning reward combines precision and recall by their harmonic mean, i.e., $F_1$ score:
\begin{equation*}
R_{\textsc{plan}} = F_1(\mathrm{precision}, \mathrm{recall}).
\end{equation*}

\paragraph{\textsc{Select} reward.} Selection is scored by exact match against the
substep's gold. Because each substep targets a single gold tool $g$
(\S\ref{sec:workflow}), the \textsc{Select} reward is
\begin{equation*}
R_{\textsc{sel}} = \mathbf{1}\bigl[\,\text{selected} = \{g\}\,\bigr],
\end{equation*}
which is $1$ only when the policy picks exactly the gold and nothing else. This
terminal reward is, per substep, the quantity the evaluation exact-match metric
measures.

\section{Experiments}
\label{sec:experiments}

Our experiments answer three questions about the trained agent.
\textbf{(i)~Success Rate.} How does a small, locally-hosted tool-selection agent
compare against inference-only baselines that drive the same task with a much
larger, general-purpose model accessed by prompting alone
(\S\ref{sec:select-results})?
\textbf{(ii)~Domain generalization.} Does an agent trained on a single tool
corpus transfer, with no retraining, to unseen corpora and domains
(\S\ref{sec:generalization})?
\textbf{(iii)~Retriever generalization.} Is the learned policy tied to the
retriever it trained on, or can the retrieval backend be swapped at test time for
a stronger, domain-specialized one (\S\ref{sec:retriever-generalization})? To answer these questions,
we train the Qwen3 model family at three model scales (1.7B, 4B, and 8B) on the ToolBench G3~\citep{qin_toolllm_2024} training split, which contains 21{,}361 user instructions over a corpus of 1{,}419 APIs for the model to select. Each user instruction in this benchmark requires multiple tools and can be decomposed into several independent substeps, so it aligns with our assumption.

\subsection{Tool Selection Success Rate}
\label{sec:select-results}
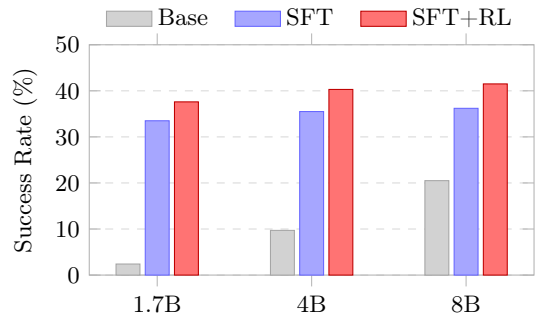
\begin{wrapfigure}{r}{0.45\textwidth}
\vspace{-\baselineskip}
\centering
{\setlength{\columnwidth}{\linewidth}\begin{tikzpicture}
\begin{axis}[
  ybar,
  bar width=9pt,
  width=\columnwidth,
  height=0.62\columnwidth,
  ymin=0, ymax=50,
  ylabel={Success Rate (\%)},
  ylabel style={font=\small},
  symbolic x coords={1.7B,4B,8B},
  xtick=data,
  xticklabel style={font=\small},
  yticklabel style={font=\small},
  ytick={0,10,20,30,40,50},
  enlarge x limits=0.22,
  ymajorgrids,
  grid style={dashed,gray!30},
  axis line style={gray!60},
  tick style={gray!60},
  legend style={at={(0.5,1.02)},anchor=south,draw=none,fill=none,font=\small,
                legend columns=3,/tikz/every even column/.append style={column sep=8pt}},
  legend cell align=left,
  area legend,
]
\addplot[fill=gray!35,draw=gray!70]  coordinates {(1.7B,2.4) (4B,9.7) (8B,20.5)};
\addplot[fill=blue!35,draw=blue!60]  coordinates {(1.7B,33.5) (4B,35.5) (8B,36.2)};
\addplot[fill=red!55,draw=red!75!black] coordinates {(1.7B,37.6) (4B,40.3) (8B,41.5)};
\legend{Base, SFT, SFT+RL}
\end{axis}
\end{tikzpicture}}
\caption{Effect of training on ToolBench G3 (test split, 2{,}124 instructions).
Success rate for Qwen3 at three scales, run through the same
\textsc{Plan}--\textsc{Retrieve}--\textsc{Select} workflow untrained (Base),
after SFT, and after SFT followed by RL (SFT+RL).}
\label{fig:g3-training}
\end{wrapfigure}
We report evaluation results on the test split of ToolBench G3, which contains 2{,}124 user instructions. Success rate is defined with an exact-match metric: tool selection on one user instruction is considered successful if and only if the selected tool set matches exactly with the filtered gold tool set of \S\ref{sec:sft}, and the success rate is averaged over all tested user instructions.
Prior work on tool retrieval mostly reports ranking metrics over the retrieved
list, such as Recall@$K$ and NDCG@$K$~\citep{qin_toolllm_2024,
quCompletenessOrientedToolRetrieval2024,shiRetrievalModelsArent2025}, or
completeness, whether every gold tool appears in the
top-$K$~\citep{quCompletenessOrientedToolRetrieval2024,shiRetrievalModelsArent2025}.
These credit a candidate list that contains the gold tools. Our metric is
stricter: the agent must commit to a set, and both a missing tool and an extra
tool count as a failure.

Figure~\ref{fig:g3-training} shows the effect of training: we compare the success rate of the untrained Qwen3 model, the model after SFT, and the model after SFT followed by RL (SFT+RL), at three scales. An untrained model performs poorly under our \textsc{Plan}--\textsc{Retrieve}--\textsc{Select} workflow. This is expected, as the model has not learned how to decompose user instructions and select tools. SFT alone raises the success rate above 30\% at every scale, and RL improves it further (e.g., from 36.2\% to 41.5\% at 8B).

We further compare our SFT+RL model
against the larger Llama-4-Maverick by prompting. We use four workflows to run Llama-4-Maverick, all with few-shot prompts and greedy decoding:
\begin{itemize}
  \item \textbf{ReAct}, a ReAct style workflow following
  ToolReAGt~\citep{braunschweilerToolReAGtToolRetrieval2025}. We use BM25 retriever and include plan decomposition to make a fair comparison with our workflow. After the same \textsc{Plan} stage as in our workflow, a ReAct loop is run for each
  substep: in every turn it writes its own search query, observes the BM25 top-$K$ candidates
  for that query, and either searches again (up to four times) or commits one tool. The union over substeps is the prediction. This differs
  from our workflow in two ways: 1) the model writes the retrieval query rather
  than reusing the substep text; 2) a substep may issue several retrievals rather
  than one.
  \item \textbf{Hierarchical search}, following
  AnyTool~\citep{duAnyToolSelfReflectiveHierarchical2024}. No retriever is needed. Instead, the
  model walks the ToolBench taxonomy from category to tool to API, and a
  self-reflection turn re-enters the tree when the selection looks incomplete.
  \item \textbf{Search-R1-style}~\citep{jin_search-r1_2025}: a single flat
  reasoning stream that interleaves \texttt{<think>}, \texttt{<search>} over
  BM25 (top-10 per search, up to six searches), and a final \texttt{<answer>}
  holding the tool set, without decomposition.
  \item \textbf{Our workflow} (\S\ref{sec:workflow}), with the untrained
  Maverick as planner and selector.
\end{itemize}

\begin{wraptable}{r}{0.45\textwidth}
\vspace{-\baselineskip}
\centering
\small
\begin{tabular}{lc}
\toprule
Method & Success Rate (\%) \\
\midrule
\multicolumn{2}{l}{\emph{Llama-4-Maverick (prompted)}} \\
\ + ReAct                        & 28.4 \\
\ + hierarchical search          & 8.6 \\
\ + Search-R1 style              & 30.6 \\
\ + our workflow                 & 10.1 \\
\midrule
\multicolumn{2}{l}{\emph{Qwen3 (SFT+RL) + our workflow}} \\
1.7B & 37.6 \\
4B   & 40.3 \\
8B   & \underline{41.5} \\
\bottomrule
\end{tabular}
\caption{Tool-selection success rate on the ToolBench G3 test split (2{,}124
instructions) under exact match: an instruction succeeds if and only if the selected tool
set equals the gold set. Llama-4-Maverick is prompted under four interaction
protocols. Qwen3 models are trained with SFT followed by RL (\S\ref{sec:method}). The best run is \underline{underlined}.}
\label{tab:g3-main}
\end{wraptable}
Table~\ref{tab:g3-main} shows that the Search-R1-style workflow pairs best with the untrained large model, reaching 30.6\% success rate. However, this result still lags behind that of our trained model. Even the smallest 1.7B model can be trained to reach 37.6\% success rate. Our workflow reaches only 10.1\% success rate for Llama-4-Maverick, which further emphasizes the necessity of training when using our workflow. Given the same Llama-4-Maverick model,
ReAct exceeds our workflow by 18\%, at the cost of up to five model calls per substep instead of one.

\subsection{Transferability across Benchmarks}
\label{sec:generalization}

A tool-selection agent is only useful if it works on corpora it was not trained on,
so we test the models trained on ToolBench G3 directly on four benchmarks. We choose
them to probe two independent axes of transferability: \emph{how far the tool corpus
and domain shift} from the training distribution, and \emph{whether the task needs a
single tool or several}. \textbf{ToolBench G2} is the mildest shift: a larger,
mostly-disjoint split of the \emph{same} RapidAPI universe (11{,}474 tools, only
$\sim$9\% shared with G3) in the same multi-tool scenario. \textbf{Gorilla}~\citep{patil_gorilla_2024} and \textbf{CRAFT}~\citep{yuan_craft_2024} come from entirely
different tool repositories and domains, exercising a much larger vocabulary shift.
All user instructions in both benchmarks have exactly one gold tool, so no \textsc{Plan} stage is needed and thus we focus on the transferability of the \textsc{Select} stage. Finally
\textbf{TaskBench}~\citep{shen_taskbench_2024} is the multi-tool out-of-domain
case: tasks from entirely different domains that each require several tools,
evaluated over each split's own tool inventory, so it is a comprehensive evaluation of both planning and selection.

\begin{table*}[t]
\centering

\small
\begin{tabular}{l@{\hspace{2em}}*{4}{>{\centering\arraybackslash}p{2.05cm}}}
\toprule
& \multicolumn{4}{c}{Success Rate (\%)} \\
\cmidrule(l){2-5}
Method & ToolBench G2 & Gorilla & CRAFT & TaskBench \\
\midrule
\emph{Llama-4-Maverick (prompted)} & & & & \\
\ + ReAct                        & 15.6 & 16.0 & 18.0 & 60.2 \\
\ + Search-R1 style              & 20.8 & 15.9 & 41.0 & \underline{80.6} \\
\ + our workflow                 & 4.4 & 9.2 & 47.9 & 36.9 \\
\midrule
\emph{Qwen3 (untrained) + our workflow} & & & & \\
1.7B & 0.9 & 5.5 & 44.8 & 5.8 \\
4B   & 4.4 & 8.7 & 49.7 & 45.6 \\
8B   & 11.2 & 8.0 & 53.4 & 55.3 \\
\midrule
\emph{Qwen3 (SFT) + our workflow} & & & & \\
1.7B & 21.8 & 17.6 & 64.7 & 54.4 \\
4B   & 23.0 & 14.5 & 65.9 & 59.2 \\
8B   & 23.9 & 26.8 & 66.4 & 66.0 \\
\midrule
\emph{Qwen3 (SFT+RL) + our workflow} & & & & \\
1.7B & 23.7 & 21.2 & 67.9 & 63.1 \\
4B   & 23.5 & 19.7 & 68.2 & 72.8 \\
8B   & \underline{25.8} & \underline{29.9} & \underline{69.0} & 72.8 \\
\bottomrule
\end{tabular}
\caption{Transferability results to new tool domains. The Qwen3 models are trained on the training split of ToolBench G3, the same checkpoints as in Table \ref{tab:g3-main}. All numbers are success rates in percent (\%). Column max \underline{underlined}. }
\label{tab:toolret-generalization}
\end{table*}

The results are shown in Table~\ref{tab:toolret-generalization}. On ToolBench G2 the trained agent transfers, though below its in-domain G3 score
($\sim$24\% vs.\ 40\% at 4B). The $8\times$ larger, mostly-disjoint corpus adds
distractors and pushes gold deeper into the BM25 ranking. Even so, it beats the
strongest \emph{untrained} baseline (Maverick + Search-R1, 20.8\%). The
same checkpoints have a clear transfer advantage on single-tool benchmarks (Gorilla and CRAFT) over Llama-4-Maverick, so the \textsc{Select} stage transfers across domains. The only exception happens on TaskBench, in which Llama-4-Maverick with the Search-R1-style workflow gives an 8\,pp higher success rate than our method. This shows that the \textsc{Plan} stage requires some domain knowledge and thus the Search-R1-style workflow, which requires no decomposition, performs better. Nonetheless, our small model still beats Llama-4-Maverick on ReAct and our workflow, which both include a \textsc{Plan} stage.

\subsection{Transferability to new retrievers}
\label{sec:retriever-generalization}

\noindent\begin{minipage}[t]{0.52\textwidth}
\vspace{0pt}
Our models are trained over BM25. The
\textsc{Retrieve} step is an interface that returns a ranked candidate set, and any
retriever that scores the corpus against a substep can fill it. We test this by
swapping BM25 for COLT~\citep{quCompletenessOrientedToolRetrieval2024}, a dense retriever \emph{trained specifically on the
ToolBench corpus}, and evaluate the models directly on the test split of ToolBench G3, without additional training.
Figure~\ref{fig:colt-retriever} reports the success rate under each retriever.

\medskip
Swapping in COLT improves the success rate at every scale. This is because COLT is an expert retriever for ToolBench, and our model inherits the stronger retriever's recall without any retraining.
We nevertheless train with BM25 because it requires no corpus-specific training, which ensures that the transfer results above do not depend on a strong in-domain retriever.
\end{minipage}%
\hfill
\begin{minipage}[t]{0.44\textwidth}
\vspace{0pt}
\centering
{\setlength{\columnwidth}{\linewidth}\begin{tikzpicture}
\begin{axis}[
  ybar,
  bar width=11pt,
  width=\columnwidth,
  height=0.62\columnwidth,
  ymin=0, ymax=50,
  ylabel={Success Rate (\%)},
  ylabel style={font=\small},
  symbolic x coords={1.7B,4B,8B},
  xtick=data,
  xticklabel style={font=\small},
  yticklabel style={font=\small},
  ytick={0,10,20,30,40,50},
  enlarge x limits=0.22,
  ymajorgrids,
  grid style={dashed,gray!30},
  axis line style={gray!60},
  tick style={gray!60},
  legend style={at={(0.5,1.02)},anchor=south,draw=none,fill=none,font=\small,
                legend columns=2,/tikz/every even column/.append style={column sep=8pt}},
  legend cell align=left,
  area legend,
]
\addplot[fill=red!55,draw=red!75!black]      coordinates {(1.7B,37.6) (4B,40.3) (8B,41.5)};
\addplot[fill=green!45!black!60,draw=green!35!black] coordinates {(1.7B,41.8) (4B,42.4) (8B,44.9)};
\legend{BM25, COLT}
\end{axis}
\end{tikzpicture}}

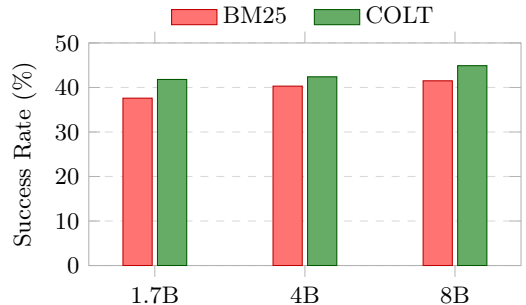
\captionof{figure}{Swapping the retriever at inference on the ToolBench G3 test split. The same G3-trained (BM25) checkpoints are run with
BM25 or COLT~\citep{quCompletenessOrientedToolRetrieval2024} behind the
\textsc{Retrieve} step, with no additional training.}
\label{fig:colt-retriever}
\end{minipage}

\subsection{Ablations}
\label{sec:ablations}

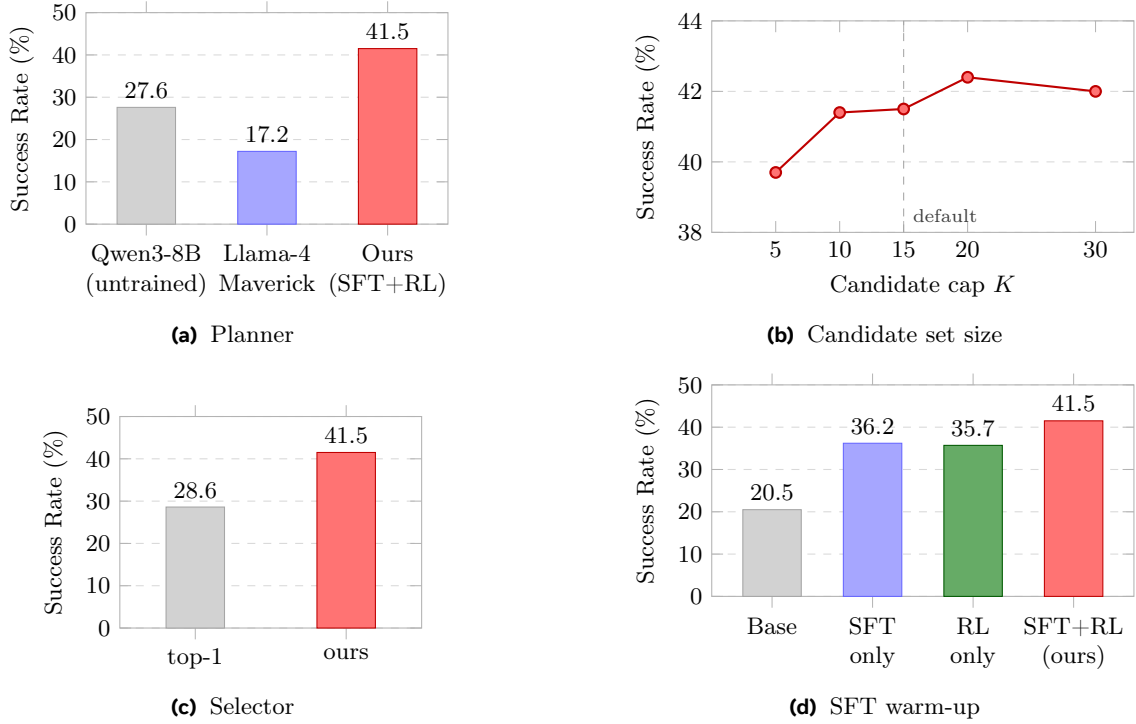
\begin{figure*}[t]
\centering
\begin{subfigure}[b]{0.48\textwidth}
\centering
{\setlength{\columnwidth}{\linewidth}\begin{tikzpicture}
\begin{axis}[
  ybar,
  bar width=22pt,
  bar shift=0pt,
  width=0.8\columnwidth,
  height=0.55\columnwidth,
  ymin=0, ymax=50,
  ylabel={Success Rate (\%)},
  ylabel style={font=\small},
  symbolic x coords={Qwen3-8B,Maverick,Ours},
  xtick={Qwen3-8B,Maverick,Ours},
  xticklabels={Qwen3-8B\\(untrained), Llama-4\\Maverick, Ours\\(SFT+RL)},
  xticklabel style={font=\small,align=center},
  yticklabel style={font=\small},
  ytick={0,10,20,30,40,50},
  enlarge x limits=0.25,
  ymajorgrids,
  grid style={dashed,gray!30},
  axis line style={gray!60},
  tick style={gray!60},
  nodes near coords,
  every node near coord/.append style={font=\small,/pgf/number format/.cd,fixed,fixed zerofill,precision=1},
]
\addplot[fill=gray!35,draw=gray!70]           coordinates {(Qwen3-8B,27.6)};
\addplot[fill=blue!35,draw=blue!60]           coordinates {(Maverick,17.2)};
\addplot[fill=red!55,draw=red!75!black]       coordinates {(Ours,41.5)};
\end{axis}
\end{tikzpicture}}
\caption{Planner}
\label{fig:planner-ablation}
\end{subfigure}
\hfill
\begin{subfigure}[b]{0.48\textwidth}
\centering
{\setlength{\columnwidth}{\linewidth}\begin{tikzpicture}
\begin{axis}[
  width=0.9\columnwidth,
  height=0.55\columnwidth,
  xmin=0, xmax=33,
  ymin=38, ymax=44,
  xlabel={Candidate cap $K$},
  ylabel={Success Rate (\%)},
  xlabel style={font=\small},
  ylabel style={font=\small},
  xtick={5,10,15,20,30},
  ytick={38,40,42,44},
  xticklabel style={font=\small},
  yticklabel style={font=\small},
  ymajorgrids,
  grid style={dashed,gray!30},
  axis line style={gray!60},
  tick style={gray!60},
]
\addplot[red!75!black,thick,mark=*,mark size=2pt,mark options={fill=red!55}]
  coordinates {(5,39.7) (10,41.4) (15,41.5) (20,42.4) (30,42.0)};
\draw[dashed,gray!70] (axis cs:15,38) -- (axis cs:15,44)
  node[pos=0.08,anchor=west,font=\scriptsize,gray!50!black] {default};
\end{axis}
\end{tikzpicture}}
\caption{Candidate set size}
\label{fig:cap-ablation}
\end{subfigure}
\\[2mm]
\begin{subfigure}[b]{0.48\textwidth}
\centering
{\setlength{\columnwidth}{\linewidth}\begin{tikzpicture}
\begin{axis}[
  ybar,
  bar width=22pt,
  bar shift=0pt,
  width=0.7\columnwidth,
  height=0.55\columnwidth,
  ymin=0, ymax=50,
  ylabel={Success Rate (\%)},
  ylabel style={font=\small},
  symbolic x coords={top1,learned},
  xtick={top1,learned},
  xticklabels={top-1, ours},
  xticklabel style={font=\small,align=center},
  yticklabel style={font=\small},
  ytick={0,10,20,30,40,50},
  enlarge x limits=0.5,
  ymajorgrids,
  grid style={dashed,gray!30},
  axis line style={gray!60},
  tick style={gray!60},
  nodes near coords,
  every node near coord/.append style={font=\small,/pgf/number format/.cd,fixed,fixed zerofill,precision=1},
]
\addplot[fill=gray!35,draw=gray!70]     coordinates {(top1,28.6)};
\addplot[fill=red!55,draw=red!75!black] coordinates {(learned,41.5)};
\end{axis}
\end{tikzpicture}}
\caption{Selector}
\label{fig:selector-ablation}
\end{subfigure}
\hfill
\begin{subfigure}[b]{0.48\textwidth}
\centering
{\setlength{\columnwidth}{\linewidth}\begin{tikzpicture}
\begin{axis}[
  ybar,
  bar width=22pt,
  bar shift=0pt,
  width=0.9\columnwidth,
  height=0.55\columnwidth,
  ymin=0, ymax=50,
  ylabel={Success Rate (\%)},
  ylabel style={font=\small},
  symbolic x coords={base,sft,rl,ours},
  xtick={base,sft,rl,ours},
  xticklabels={Base, SFT\\only, RL\\only, SFT+RL\\(ours)},
  xticklabel style={font=\small,align=center},
  yticklabel style={font=\small},
  ytick={0,10,20,30,40,50},
  enlarge x limits=0.2,
  ymajorgrids,
  grid style={dashed,gray!30},
  axis line style={gray!60},
  tick style={gray!60},
  nodes near coords,
  every node near coord/.append style={font=\small,/pgf/number format/.cd,fixed,fixed zerofill,precision=1},
]
\addplot[fill=gray!35,draw=gray!70]     coordinates {(base,20.5)};
\addplot[fill=blue!35,draw=blue!60]     coordinates {(sft,36.2)};
\addplot[fill=green!45!black!60,draw=green!35!black] coordinates {(rl,35.7)};
\addplot[fill=red!55,draw=red!75!black] coordinates {(ours,41.5)};
\end{axis}
\end{tikzpicture}}
\caption{SFT warm-up}
\label{fig:warmup-ablation}
\end{subfigure}
\caption{Ablations on the ToolBench G3 test split with the 8B checkpoint.
(a) We ablate the \textsc{Plan} stage by replacing the planner model with untrained Qwen3-8B and Llama-4-Maverick.
(b)~Success rate as the number of retrieved candidates $K$ handed to
\textsc{Select} varies. The dashed line marks the default $K=15$.
(c)~The learned \textsc{Select} stage (``ours'') is replaced by a fixed rule
that commits the BM25 top-1 candidate of each substep (``top-1'').
(d) We run RL directly from the untrained model (RL only) to ablate SFT.}
\label{fig:ablations}
\end{figure*}

The ablation studies are all evaluated on the test split of ToolBench G3.

\noindent \emph{Planner.} We ablate the model used for the \textsc{Plan} stage, by replacing the trained 8B model with: 1) the untrained
Qwen3-8B; 2) Llama-4-Maverick. We hold the \textsc{Retrieve} and \textsc{Select} stages fixed. As shown in Figure~\ref{fig:planner-ablation}, the
trained planner is the best of the three. The much larger Maverick planner performs \emph{worst}, indicating that training, not model scale, drives planning quality on this task.

\noindent \emph{Candidate set size.} We sweep the retrieved candidate set size $K$, i.e., the number of candidates handed to the \textsc{Select} stage, on the fixed 8B checkpoint. Success rate increases with $K$ up to $K=20$ and does not improve further at $K=30$ (Figure~\ref{fig:cap-ablation}): a larger set is more likely to contain the gold tool, but additional candidates also act as distractors. We set $K=15$ as our training and evaluation configuration to balance context consumption and performance.

\noindent \emph{Selector.} We replace the \textsc{Select} stage with a fixed selection
rule that commits the retriever's top-1 tool for each substep. Against the
\emph{same} BM25 retriever the pipeline uses, the learned selector improves success by about 13\% over taking the top-1 candidate (Figure~\ref{fig:selector-ablation}), because the highest-scoring BM25 candidate is often not the tool the substep requires.

\noindent \emph{SFT warm-up.} To test whether the SFT stage is necessary, we
run our RL recipe \emph{directly} from the off-the-shelf 8B model, with no SFT
initialization, which is viable because the base already completes the workflow with a
non-zero success rate for GRPO to bootstrap from. Direct RL lifts the base model performance from 20.5\% to 35.7\%, but lands at a similar level to SFT only (36.2\%) and
below the full SFT+RL recipe (41.5\%), supporting the value of SFT as a warm-up procedure
(Figure~\ref{fig:warmup-ablation}).

\FloatBarrier

\section{Conclusion}
\label{sec:conclusion}

Tool selection over a large tool corpus is a recurring sub-routine of agentic
tool calling, and prior work runs it by prompting proprietary models to
operate a retriever. We showed that this capability can be \emph{trained}
into a small open model. Under a minimal
\textsc{Plan}--\textsc{Retrieve}--\textsc{Select} workflow, supervised
warm-up on teacher-written trajectories followed by RL with a
retrieval-based planning reward and an exact-match selection reward lifts Qwen3 models of 1.7B to 8B parameters past
Llama-4-Maverick (402B) on ToolBench G3, with no human annotation beyond the
gold tool sets. Because the skills involved, i.e., instruction decomposition and
tool selection from description, are not tied to a specific tool corpus, the same
checkpoints transfer without retraining to new tool domains and to new retrievers, inheriting the
stronger retriever's recall. We hope this work encourages further study of tool selection with small LLMs.

\section*{Acknowledgments}
We thank Tal Lancewicki for discussion and valuable feedback during the project. This work was partially supported by the National Science Foundation under Grant CCF-2106778.

\clearpage
\newpage
\bibliographystyle{assets/plainnat}
\bibliography{meta_paper}

\clearpage
\newpage
\beginappendix

\section{Experiment Details}
\label{app:exp-details}

\subsection{Models and Benchmarks}
\label{app:models-benchmarks}

Links to all models, benchmarks, and software used in this paper are collected
in Appendix~\ref{app:links} (Table~\ref{tab:links}).

\paragraph{Trained models.} We train Qwen3-1.7B, Qwen3-4B, and
Qwen3-8B. Qwen3 can emit a \texttt{<think>} block before its
answer, and our trained models use this block for the brief reasoning that
precedes each JSON action: the SFT targets place the teacher-written rationale
inside \texttt{<think>}\ldots\texttt{</think>} followed by the action
(Appendix~\ref{app:prompts}), and the SFT and SFT+RL models are run with
thinking enabled during SFT, RL, and evaluation. The reasoning they produce is
therefore the short, task-specific rationale learned in SFT and shaped by RL,
not Qwen3's open-ended native deliberation. The \emph{untrained} Qwen3 models
(the Base rows of \S\ref{sec:select-results}, the untrained planner of
\S\ref{sec:ablations}, and the RL-only ablation, which starts from the base
model) are instead run with thinking disabled
(\texttt{enable\_thinking=False}), because the base model's native thinking
block routinely exhausted the generation budget before producing an action. A
single model serves both the \textsc{Plan} and \textsc{Select} stages.

\paragraph{Teacher and baseline model.} Llama-4-Maverick is a mixture-of-experts model
with 128 experts, 17B active parameters, and 402B total parameters, so it
is roughly two orders of magnitude larger than our largest trained model in
total parameters and twice as large in active parameters. It plays two roles. It
is the teacher for gold filtering, plan decomposition, and reasoning generation
in the data curation pipeline of \S\ref{sec:sft}, and it is the prompted model
behind all four inference-only baselines of \S\ref{sec:select-results}.

\paragraph{Retrievers.} The default retriever is BM25 implemented with the
\texttt{bm25s} library~\citep{lu2024bm25s} (Lucene scoring variant) using $k_1=1.8$ and $b=1.0$,
English stopword removal, and no stemming. Each tool is indexed by its
serialized documentation entry (tool and API name, description, and parameter
fields). Candidates are kept if their score exceeds $\tau$ and then truncated
to the top $K$ ($\tau=3$, $K=15$ unless stated otherwise in
Table~\ref{tab:bench-stats}). The same configuration is used for the planning
reward, for SFT data generation, and at evaluation. For the retriever-swap
experiment (\S\ref{sec:retriever-generalization}) we use the public COLT
checkpoint for G3~\citep{quCompletenessOrientedToolRetrieval2024}, a Contriever
encoder~\citep{izacard2021unsupervised} trained
on the ToolBench G3 training split, together with its graph-propagated tool
embeddings; the retriever returns dot-product similarity scores.

\paragraph{Benchmarks.} Table~\ref{tab:bench-stats} summarizes the corpora.
\textbf{ToolBench G3}~\citep{qin_toolllm_2024} is the training benchmark: its
corpus has 1{,}419 RapidAPI tools, the training split has 21{,}361 user
instructions, and the test split has 2{,}124. Every instruction requires multiple
tools and decomposes into several independent substeps. All training, the main
results, the ablations, and the retriever-swap experiment use G3.
\textbf{ToolBench G2} is a larger, mostly disjoint split of the same RapidAPI
universe (11{,}474 tools, about 9\% shared with G3) in the same multi-tool setting;
we evaluate the 7{,}238 test instructions that survive the same gold filtering
as G3 (\S\ref{sec:sft}).
\textbf{Gorilla}~\citep{patil_gorilla_2024} covers tool calls to three
machine-learning model hubs, with one source split each
(\texttt{huggingface}, \texttt{pytorch}, \texttt{tensor}; 500, 43, and 55
instructions). \textbf{CRAFT}~\citep{yuan_craft_2024} covers tool use for math,
tabular, and visual question answering (\texttt{math-algebra}, \texttt{tabmwp},
\texttt{vqa}; 280, 174, and 200 instructions). Every Gorilla and CRAFT
instruction has exactly one gold tool. \textbf{TaskBench}~\citep{shen_taskbench_2024}
is the multi-tool out-of-domain case: three source splits (\texttt{daily},
\texttt{multimedia}, \texttt{huggingface}; 103 instructions in total), each with
its own small inventory of 23--40 tools, where each instruction requires several
tools (2.6 on average). ToolBench G2 and G3 use the retrieval splits released
with ToolBench. Gorilla, CRAFT, and TaskBench use the packaging of
ToolRet~\citep{shiRetrievalModelsArent2025}, which provides instruction-framed
queries and a unified corpus format; we restrict each split's corpus to its own
source's tool inventory rather than ToolRet's merged corpus, matching the
original benchmarks. All user instructions and tool documentation in every
benchmark are in English. The G3 test split has 2{,}373 instructions, of which
2{,}124 (89.5\%) survive the gold filtering of \S\ref{sec:sft}; the G2 test split
has 8{,}250, of which 7{,}238 (87.7\%) survive. Gorilla, CRAFT, and TaskBench are
evaluated on their original gold labels without filtering.

\begin{table}[h]
\centering
\footnotesize
\setlength{\tabcolsep}{3pt}
\begin{tabular}{lrrccc}
\toprule
Benchmark & Instr. & tools & Gold & $\tau$ & $K$ \\
\midrule
ToolBench G3 (test) & 2{,}124 & 1{,}419 & multi & 3 & 15 \\
ToolBench G2 (test) & 7{,}238 & 11{,}474 & multi & 3 & 15 \\
Gorilla (3 splits)  & 598 & 43--907$^\dagger$ & 1 & 3 & 20 \\
CRAFT (3 splits)    & 654 & 180--525$^\dagger$ & 1 & 2 & 20 \\
TaskBench (3 splits)& 103 & 23--40$^\dagger$ & multi & 0 & 15 \\
\bottomrule
\end{tabular}
\caption{Evaluation benchmarks. ``Instr.'' is the number of evaluated
instructions, ``tools'' the size of the corpus retrieved over, ``Gold'' the
number of gold tools per instruction, and $\tau$ and $K$ the BM25 score
threshold and candidate cap. About 9\% of the G2 tools also appear in G3.
$^\dagger$Range over the source splits, each retrieved over separately.}
\label{tab:bench-stats}
\end{table}

\subsection{Training}
\label{app:train}

\paragraph{Training data.} All training uses the ToolBench
G3~\citep{qin_toolllm_2024} training split only. The split has 21{,}361
instructions with 57{,}836 (instruction, gold tool) pairs, i.e., 2.71 gold tools
per instruction. The gold tool filtering of \S\ref{sec:sft} keeps 54{,}695 pairs
(94.6\%); the plan decomposition step then drops a further 1{,}928 gold tools
that the teacher judges unnecessary for the instruction. In total 1{,}268
instructions (5.9\%) are discarded, either because no gold tool survives or
because the teacher fails the output validation three times, leaving 20{,}093
curated training instructions. The SFT corpus built from them contains
19{,}841 \textsc{Plan} examples and 38{,}293 \textsc{Select} examples, each a
single (prompt, reasoning + action) pair. A held-out slice of 119 training
instructions (with its 783 SFT examples) serves as a validation set during
development. The RL prompt set is derived from the same curated instructions:
one \textsc{Plan} prompt per instruction and one \textsc{Select} prompt per
teacher-written substep.

\paragraph{Supervised fine-tuning.} We fine-tune all parameters with
LLaMA-Factory~\citep{zheng2024llamafactory} using DeepSpeed
ZeRO-3~\citep{rajbhandari2020zero} in bf16. Hyperparameters are listed in
Table~\ref{tab:hparams}.

\paragraph{Reinforcement learning.} We use GRPO~\citep{shao_deepseekmath_2024}
as implemented in VeRL~\citep{sheng2025hybridflow}, with rollouts generated by
vLLM~\citep{kwon2023efficient}, starting from the SFT checkpoint. Hyperparameters are listed in
Table~\ref{tab:hparams}, and are identical across the three model scales.

\paragraph{Software versions.} Training uses VeRL at commit \texttt{e52747a4}
(v0.8.0 plus 147 commits), vLLM 0.11.0, PyTorch 2.8.0 with CUDA 12.8,
Transformers 4.56.1, DeepSpeed 0.18.4, and FlashAttention 2.8.1. Evaluation
uses a separate environment with vLLM 0.22.1, PyTorch 2.11.0, Transformers
5.10.2, \texttt{bm25s} 0.3.9, and \texttt{sentence-transformers} 5.5.1 for
the COLT encoder. All reported numbers are produced with the evaluation
environment.

\begin{table}[h]
\centering
\footnotesize
\setlength{\tabcolsep}{2pt}
\begin{tabular}{lcc}
\toprule
 & SFT & RL (GRPO) \\
\midrule
Framework                   & LLaMA-Factory & VeRL + vLLM \\
Trainable parameters        & full & full \\
Learning rate (AdamW)       & $10^{-5}$ & $10^{-6}$ \\
LR schedule                 & cosine & constant \\
Epochs or steps             & 3 epochs$^\ddagger$ & 300 steps \\
Effective batch size        & 128 sequences & 64 prompts \\
Rollouts per prompt         & -- & 16 \\
Sampling temperature        & -- & 0.9 \\
KL coefficient              & -- & 0.001 \\
Updates per batch           & -- & 1 \\
Max prompt / completion     & 4{,}096 (packed) & 4{,}096 / 4{,}096 \\
Hardware                    & 8$\times$A100 & 8$\times$A100 \\
\bottomrule
\end{tabular}
\caption{Training hyperparameters. The same values are used at all three model
scales. The SFT schedule has a 3\% linear
warm-up. }
\label{tab:hparams}
\end{table}

\subsection{Evaluation}
\label{app:eval}

\paragraph{Metric.} All results report the exact-match success rate of
\S\ref{sec:select-results}: an instruction counts as a success if and only if the
union of selected tool IDs over all substeps equals the filtered gold tool set. A
missing tool, an extra tool, or an unparseable action all count as failures. All models are decoded greedily (temperature~0), with at most
2{,}048 new tokens per call and at most 8 substeps per plan. All numbers are
from a single run; repeated runs of the same checkpoint differ by one to two
instructions because of non-deterministic batched inference.

\paragraph{ToolBench G3 evaluation.} The main results (Table~\ref{tab:g3-main},
Figure~\ref{fig:g3-training}) are on the G3 test split over the full G3 corpus
(Table~\ref{tab:bench-stats}), with BM25, $\tau=3$, and $K=15$ as in training.
The ablations of \S\ref{sec:ablations} use the same split and the 8B
SFT+RL checkpoint; the candidate-set sweep (Figure~\ref{fig:cap-ablation})
evaluates $K\in\{5,10,15,20,30\}$ with everything else fixed.

\paragraph{Llama-4-Maverick baselines.} All four protocols of
\S\ref{sec:select-results} prompt the same Llama-4-Maverick
(Appendix~\ref{app:models-benchmarks}) with greedy decoding and at most 1{,}024
new tokens per call (2{,}048 for our workflow).

\emph{ReAct} first decomposes the
instruction into at most five substeps with a prompt that includes two few-shot
decomposition examples, then runs one ReAct loop per substep over BM25: Each turn
the model sees the substep, its search history, and the top-10 results of its
latest search, and chooses to search again, select one tool, or skip; the loop
may issue at most four searches before committing.

\emph{Hierarchical search}
follows AnyTool~\citep{duAnyToolSelfReflectiveHierarchical2024}: the model
descends the ToolBench category $\to$ tool $\to$ API taxonomy (G3: 36
categories and 241 tools; G2: 49 categories and 3{,}391 tools), choosing at most
3 categories, 6 tools per category, and 15 APIs per tool, with API descriptions
truncated to 200 characters, and re-enters the tree after one self-reflection
round.

\emph{Search-R1-style} follows \citet{jin_search-r1_2025}: a single
generation interleaving \texttt{<think>}, \texttt{<search>} (BM25 top-10 per
search, at most six searches), and a final \texttt{<answer>} containing the tool
set, with no decomposition; the prompt includes one three-turn worked example,
and generation is stopped at each \texttt{</search>} tag so the model cannot
fabricate retrieval results.

\emph{Our workflow} runs the untrained Maverick as
both planner and selector with the prompts of Appendix~\ref{app:prompts} and no
few-shot examples.

\paragraph{Transfer benchmark evaluations.} Retrieval settings per
benchmark are given in Table~\ref{tab:bench-stats}.

\paragraph{Retriever swap.} For the COLT experiment
(\S\ref{sec:retriever-generalization}) we replace BM25 scores with COLT
similarity scores (Appendix~\ref{app:models-benchmarks}) inside the
\textsc{Retrieve} stage. Because COLT scores are on a different scale from
BM25, the threshold is set to $\tau=0.6$; $K=15$ is unchanged. The model
checkpoints and prompts are identical to the BM25 runs.

\section{Links to Models, Benchmarks, and Software}
\label{app:links}

Table~\ref{tab:links} lists the public location and license of every model
checkpoint, benchmark, and software framework used in this paper.

\begin{table*}[t]
\centering
\scriptsize
\setlength{\tabcolsep}{3pt}
\resizebox{\textwidth}{!}{%
\begin{tabular}{lll}
\toprule
Resource & URL & License \\
\midrule
Qwen3-1.7B~\citep{yang2025qwen3} & \url{https://huggingface.co/Qwen/Qwen3-1.7B} & Apache-2.0 \\
Qwen3-4B~\citep{yang2025qwen3} & \url{https://huggingface.co/Qwen/Qwen3-4B} & Apache-2.0 \\
Qwen3-8B~\citep{yang2025qwen3} & \url{https://huggingface.co/Qwen/Qwen3-8B} & Apache-2.0 \\
Llama-4-Maverick~\citep{meta2025llama4} & \url{https://huggingface.co/meta-llama/Llama-4-Maverick-17B-128E-Instruct} & Llama 4 Community License \\
COLT checkpoint (G3)~\citep{quCompletenessOrientedToolRetrieval2024} & \url{https://huggingface.co/Tool-COLT/contriever-base-msmarco-v1-ToolBenchG3} & Unspecified \\
BM25 (\texttt{bm25s})~\citep{lu2024bm25s} & \url{https://github.com/xhluca/bm25s} & MIT \\
\midrule
ToolBench (G2, G3)~\citep{qin_toolllm_2024} & \url{https://github.com/OpenBMB/ToolBench} & Apache-2.0 \\
Gorilla~\citep{patil_gorilla_2024} & \url{https://github.com/ShishirPatil/gorilla} & Apache-2.0 \\
CRAFT~\citep{yuan_craft_2024} & \url{https://github.com/lifan-yuan/CRAFT} & Unspecified \\
TaskBench~\citep{shen_taskbench_2024} & \url{https://github.com/microsoft/JARVIS/tree/main/taskbench} & MIT \\
ToolRet packaging~\citep{shiRetrievalModelsArent2025} & \url{https://github.com/mangopy/tool-retrieval-benchmark} & Apache-2.0 \\
\midrule
LLaMA-Factory (SFT)~\citep{zheng2024llamafactory} & \url{https://github.com/hiyouga/LLaMA-Factory} & Apache-2.0 \\
VeRL (RL)~\citep{sheng2025hybridflow} & \url{https://github.com/volcengine/verl} & Apache-2.0 \\
vLLM (inference)~\citep{kwon2023efficient} & \url{https://github.com/vllm-project/vllm} & Apache-2.0 \\
\bottomrule
\end{tabular}}
\caption{Public locations and licenses of the resources used in this paper.
Top: models and retrievers. Middle: benchmarks. Bottom: software.
``Unspecified'' means that no license is declared at the public location; these resources are used for research evaluation only.}
\label{tab:links}
\end{table*}

\section{Artifact Use and Release}
\label{app:artifact-use}

\paragraph{Use of existing artifacts.} All models, benchmarks, and software in
Table~\ref{tab:links} are used for non-commercial research, consistent with
their licenses and, where stated, their intended use. The benchmarks are used
for the purpose they were released for, namely evaluating tool retrieval and
selection. Llama-4-Maverick is used as a teacher to generate the SFT
supervision of \S\ref{sec:sft} and as a prompted baseline, under the Llama 4
Community License. The COLT checkpoint is used only for the retriever-swap
evaluation of \S\ref{sec:retriever-generalization}. Where a resource declares
no license (Table~\ref{tab:links}), we use it only for research evaluation and
do not redistribute it. All data are English tool documentation and synthetic
user instructions created by the benchmark authors; we did not collect any new
data from people.

\paragraph{Created artifacts.} This work creates three derived artifacts: the
curated gold tool sets and teacher-written SFT trajectories for ToolBench G3 and
G2 (\S\ref{sec:sft}), the trained Qwen3 checkpoints, and the training and
evaluation code. Because the SFT data are generated with Llama-4-Maverick, the
trained checkpoints are derivatives of Llama outputs. None of these artifacts
are released, and they are used only for the experiments
reported here. The intended use of the trained models is research on tool
selection; they are not validated for deployment in agents that execute
consequential actions. All information needed to
re-implement the pipeline, including prompts (Appendix~\ref{app:prompts}),
hyperparameters (Table~\ref{tab:hparams}), software versions
(Appendix~\ref{app:train}), and data statistics, is reported in this paper.

\section{Prompt Templates}
\label{app:prompts}

For completeness we list the exact prompts used for each stage of
the evaluation workflow (\S\ref{sec:workflow}). Every LLM
call is stateless: it consists of one system prompt and one user message,
re-rendered from structured state, so the model never sees a growing transcript.
Placeholders in \texttt{\{braces\}} are filled in by the workflow at render time.

\paragraph{\textsc{Plan}.} A single call decomposes the user task into an ordered list of
substeps.

\begin{promptbox}{PLAN: System prompt}
\begin{Verbatim}[breaklines=true,breaksymbolleft={},breakindent=0pt,fontsize=\small]
You are a tool-search agent in the PLAN phase. Decompose the user
task into an ordered list of sub-steps.

Output: brief reasoning, then a JSON object
{"name": "PLAN", "substeps": [{"text": "..."}, ...]}
\end{Verbatim}
\end{promptbox}

\begin{promptbox}{PLAN: User prompt}
\begin{Verbatim}[breaklines=true,breaksymbolleft={},breakindent=0pt,fontsize=\small]
User task: {query}
\end{Verbatim}
\end{promptbox}

\paragraph{\textsc{Select}.} A single call over the $\leq K$ retrieved candidates, the
only stage in which full tool documentation enters the context. The optional
leading \texttt{User task} line restores query detail that the decomposition may
have dropped.

\begin{promptbox}{SELECT: System prompt}
\begin{Verbatim}[breaklines=true,breaksymbolleft={},breakindent=0pt,fontsize=\small]
You are a tool-search agent in the SELECT phase. Pick id(s) from
the candidate list that solve the sub-step.

Output: brief reasoning, then a JSON object
{"name": "SELECT", "ids": ["<id>", ...]}
\end{Verbatim}
\end{promptbox}

\begin{promptbox}{SELECT: User prompt}
\begin{Verbatim}[breaklines=true,breaksymbolleft={},breakindent=0pt,fontsize=\small]
User task: {query}

Sub-step: {substep}

Candidates ({n}):
[{id_1}]
{doc_1}

[{id_2}]
{doc_2}
...
\end{Verbatim}
\end{promptbox}

\end{document}